\documentclass[final,3p,twocolumn]{elsarticle}

\usepackage{lineno,hyperref}
\modulolinenumbers[5]

\usepackage{lipsum}
\usepackage{booktabs}
\usepackage[capitalise,nameinlink,noabbrev]{cleveref}
\usepackage{csquotes}
\usepackage{rotating}
\usepackage{makecell}
\usepackage{float}
\usepackage{xcolor}

\newcommand{\sgrYes}{X}

\newcommand{\sgrROT}[1]{
\begin{sideways}
	\footnotesize{#1}
\end{sideways}
}

\journal{an Elsevier Journal}

\begin{document}

\begin{frontmatter}

\title{Software Architecture for Next-Generation AI Planning Systems}

\author[swt_stud_address]{Sebastian Graef\corref{mycorrespondingauthor}}
\ead{sebastian@graef.me}
\cortext[mycorrespondingauthor]{Corresponding author}

\author[IAAS_address]{Ilche Georgievski}
\ead{ilche.georgievski@iaas.uni-stuttgart.de}

\address[swt_stud_address]{M.Sc. in Software Engineering, University of Stuttgart, Germany}
\address[IAAS_address]{Service Computing Department, Institute of Architecture of Application Systems, University of Stuttgart, Germany}

\begin{abstract}
Artificial Intelligence (AI) planning is a flourishing research and development discipline that provides powerful tools for searching a course of action that achieves some user goal. While these planning tools show excellent performance on benchmark planning problems, they represent challenging software systems when it comes to their use and integration in real-world applications. In fact, even in-depth understanding of their internal mechanisms does not guarantee that one can successfully set up, use and manipulate existing planning tools. We contribute toward alleviating this situation by proposing a service-oriented planning architecture to be at the core of the ability to design, develop and use next-generation AI planning systems. We collect and classify common planning capabilities to form the building blocks of the planning architecture. We incorporate software design principles and patterns into the architecture to allow for usability, interoperability and reusability of the planning capabilities. Our  prototype planning system demonstrates the potential of our approach for rapid prototyping and flexibility of system composition. Finally, we provide insight into the qualitative advantages of our approach when compared to a typical planning tool. 
\end{abstract}

\begin{keyword}
AI Planning \sep Software Architecture \sep Planning Capabilities \sep SOA
\end{keyword}

\end{frontmatter}


\section{Introduction} 
Artificial Intelligence (AI) planning computationally solves the problem of finding a course of action that achieve some user goal. The planning problem is usually based on a model that describes the world in a given domain and how to change such a world. The course of action that solves the the problem is called a plan. Different planning techniques have been developed that produce solution plans. In addition to solving techniques, the AI planning community has invested a lot of effort on approaches for addressing other planning-related tasks, such as representation and compilation of domain knowledge, validation of plans, monitoring and execution of plans, and visualisation and explanation of plans and decisions made. We refer to capacities as these as \textit{planning capabilities}. 

AI planning research is at a stage where software tools, called planners, can produce plans consisting of hundreds of steps in basically no time. The planners are traditionally evaluated on benchmark domains the majority of which are synthetically created for the International Planning Competitions (IPCs), and only few seem to stem directly from the real world (e.g,. power supply restoration~\cite{thiebaux2001:supply-restoration}).

The excellence of planners is largely demonstrated within the scope of scientific applications. Outside of this scientific scope, there are several studies reporting on the use of planners for real-world domains, and on the integration of planning in actual applications
and systems. Examples include Web service composition~\cite{kaldeli2011:continual}, games~\cite{orkin2003:games}, space missions~\cite{muscettola1998:spacecraft}, and ubiquitous computing~\cite{georgievski2016:puc}. Except maybe for the domain of games, there is hardly any other evidence on the wide and continuous use of planning in real-world applications.

Given that AI planning has been recognised as capable of application to realistic domains, the question that arises is why AI planning has not been widely used in practice. It might be that real-world and modern applications require planning tools that are interoperable, modular, portable, maintainable, and reusable. Planning software components need to be compatible with each other and other application components, thus designed for interoperability. Modularity requires from planning tools to have well-defined and independent functionality. The modularity together with evolvability should provide for high maintainability, i.e., easy fixing of issues or modifying some capability. 

Software systems with such considerations are typically designed and developed following fundamental software design principles at both architectural and component level. Research in AI planning appears to alienate the planning tools from such considerations principally due to their orthogonality to knowledge representation and reasoning as primary concerns of the community. By examining existing planning tools, one can notice that software design decisions are neither known nor considered. For example, the influential Fast Forward (FF) planner is associated with an architecture that consists of two techniques rather than components or separate concerns~\cite{Hoffmann2001}, and another prominent planner, the Fast Downward (FD) planner, is associated with a diagram representing three phases that sort of give a coarse separation of concerns~\cite{helmert-jair2006}. Similar observations can be made about planning tools for domain parsing~\cite{upddl-parser}, domain modelling~\cite{vaquero2013:itsimple}, plan validation~\cite{howey2004:val}, and so on.

So, there are no specific engineering approaches that capture the peculiarities and design decisions of current planning tools. Also, apart from standard languages for modelling planning problems, there are no standards for planning capabilities. A designer or developer must get bogged down in theory and detail of source codes to use and integrate planning capabilities. This situation gets even worse when the developer would need to look at the scale and maintenance of planning capabilities.  

Our aim is to address usability, interoperability and reusability by looking at a collection of planning capabilities through the prism of Service-Oriented Architecture (SOA)~\cite{papazoglou2003:soc}. We summarise our contributions as follows.

\begin{itemize}
     \item We reduce planning complexity by providing a set of flatting and sizing planning capabilities. This collection of capabilities is not only beneficial to our approach but it can also serve as a stepping stone for future software designs of AI planning systems.
     \item We propose a novel classification of planning capabilities from an operational and technical view. These classifications provide a new perspective over planning capabilities and new insights about their properties, commonalities and differences. 
     \item We propose a new planning architecture in which planning capabilities are designed as loosely coupled services that communicate via messaging. This service-oriented architecture is the first one that considers software design principles and patterns for the purpose of providing usable, interoperable and reusable planning capabilities. Another important contribution of the planning architecture is that it separates technical issues from planning-related challenges.
     \item We implement a prototype planning system and demonstrate the potential of rapid prototyping, the flexibility of integrating different service implementations, and the possibility to compose the planning capabilities depending on the application needs.
     \item We asses the quality of our approach and show that our approach has the potential to improve usability, interoperability and reusability of planning capabilities in comparison to typical existing planning systems.
\end{itemize}

To the best of our knowledge, no approach exists that discerns a collection of planning capabilities from existing planning tools upon which service orientation together with software design principles and patterns are applied. Our work is the first significant step towards placing a planning architecture at the core of ability to design, develop and use next-generation AI planning system for real-world, modern and future applications.

The remainder of the article is organised as follows. Section~\ref{sec:bg} briefly introduces the field of AI planning and SOA. Section~\ref{sec:rw} provides a discussion on related work. Section~\ref{sec:pc} presents relevant AI planning capabilities and their classification. Section~\ref{sec:design} introduces the proposed design of planning architecture. Section~\ref{sec:impl} gives insights into the prototype planning system. Section~\ref{sec:benefits} presents a qualitative assessment of our proposal. Section~\ref{sec:discussion} provides a discussion on our architecture, prototype and assessment. Section~\ref{sec:conclusion} finalises the article with concluding remarks.

\section{Background}\label{sec:bg}
We introduce the field of AI planning, including the basic concepts related to modelling and solving planning problems, languages for specifying planning problems, and planners used to solve planning problems. After this introduction, we turn to describing service-oriented architectures with an emphasis on loose coupling and integration via messaging and patterns.

\subsection{AI Planning}
AI planning is a growing research and development discipline that originated around 1966 from the need to give autonomy to Shakey, the first general-purpose mobile robot~\cite{nilsson1984:shakey}. Over time, many planning approaches have been proposed for addressing general and more specific problems and aspects. A \textit{planning problem}, which consists of an initial state, a goal state and a set of actions, is solved by searching for actions that lead from the initial state to the goal state. The initial state describes how the world looks like at the moment the planning process starts. The goal state describes the goals of the user. Actions represent domain knowledge and describe the transitions from one state to another. An action consists of preconditions, which are conditions that must be satisfied in the current state so that the action can be applied, and effects, which are conditions that must hold in the state after the application of the action. A \textit{plan} as a structure of actions is a solution to a planning problem if the preconditions of the plan's first action are satisfied in the initial state and if the goal is satisfied after the execution of the plan.

Planning approaches often rely on several simplifying assumptions, such as completeness and full observability of the initial state, determinism of actions, modifiability of states only by planning actions, hard goals, sequentiality of plans, and no explicit use of time~\cite{ghallab2004:automated}. These assumptions are too restrictive for real-world domains, where completeness of information cannot be assured and plan execution cannot be guaranteed to succeed as expected during planning. In addition, for some domains, actions and goal states may not be sufficient or adequate to express complex or structured domain knowledge (e.g., as hierarchies). So, much work has focused on developing planning approaches that relax one or more of the assumptions (e.g.,~\cite{kaldeli2016:planning,bertoli2001:nondeterministic,kaldeli2009:extended,mausam2008:durative}), and support specifying additional knowledge, such as Hierarchical Task Networks (HTN) planning~\cite{georgievski2015:htn}.

AI planning approaches are typically implemented as software tools, called \textit{planners}, with the purpose to demonstrate their performance in terms of speed of planning and quality of plans. Planners are often only solvers of planning problems, and in some cases planners may include other functionality, such as modelling and plan execution. Among the most prominent ones are FF and FD due to their performance advantages. These planners represent the base implementation for many other planners (e.g., TFD~\cite{eyerich2009:tfd}, PROBE~\cite{lipovetzky2011:probe}). Among HTN planners, the Simple Hierarchical Ordered Planner (SHOP)~\cite{nau1999:shop} is the most well-known planner that has been extended with numerous additional functionality, such as partially ordered tasks in SHOP2~\cite{nau2003shop2}, preferences in HTNPlan-P~\cite{sohrabi2009:htnplan-p}, etc. Other hierarchical planners include SIADEX~\cite{fdez2006bringing}, SH~\cite{georgievski2017:pmc}, and PANDA~\cite{holler2021:panda}.

Planning problems are typically specified in some syntax. The most common specification syntax is the Planning Domain Definition Language (PDDL). PDDL is a Lisp-based language developed for the needs of the first IPC in 1998~\cite{mcdermott1998:pddl}, and has become a defacto standard language for specifying planning problems also outside IPC. To support modelling more realistic planning problems, PDDL was later extended with additional constructs, such as numeric fluents, plan metrics and durative actions~\cite{fox2003:pddl21}, preferences~\cite{gerevini2005:pddl3}, probabilistic effects~\cite{younes2004:ppddl} and mixed discrete-continuous dynamics~\cite{fox2006:pddl+}. For modelling planning domains with HTNs, a few languages exist, such as SHOP2-based syntax, Hierarchical Planning Definition Language (HPDL), which is the first language for hierarchical planning built on the basis of PDDL~\cite{georgievski2013:hpdl}, and Hierarchical Domain Definition Language (HDDL)~\cite{hoeller2020:hddl}, a merely slight modification of HPDL.  

\subsection{Service-Oriented Architecture}\label{sec:bg:soc-soa}
SOA defines a way to make software components interoperable and reusable. These  can be accomplished by using common communication standards and design patterns in such a way that software components can be integrated in existing and new systems without the need of deep integration. 

The basic building block of an SOA is a service, which encapsulates a capability that addresses a specific functional requirement, hiding away code and data integration details required for the service execution. This definition entails service loose coupling, a key feature of SOA. Loose coupling ensures that services can be invoked with little or no knowledge of how the integration is achieved in the implementation. That is, loose coupling intentionally sacrifices interface optimisation to achieve flexible interoperability among systems that are disparate in technology, location, performance, and availability~\cite{kayeLooselyCoupled}. In other words, as services are typically distributed, the communication between them may be affected due to various reasons, such as network outage, failure of interacting services, speed of data processing or computation, and provisioning of resources. Therefore, a set of autonomies of loose coupling among services should be enabled, namely reference autonomy, time autonomy, format autonomy, and platform autonomy~\cite{Fehling2014}. Reference autonomy indicates that interacting components do not know each other. Only the place where to read/write data is known. Time autonomy means the communication between interacting components is asynchronous. Components access channels at their own pace and the data exchanged is persistent. Format autonomy allows components to exchange data in different formats, and each component may not know the data format of the interacting component. Platform autonomy indicates that interacting components may not be implemented in the same programming language and run on the same environment.

Software components can work together and exchange information while maintaining loose coupling by using messaging. Under the messaging integration style, two components exchange a message via a message channel~\cite{hohpe2004enterprise}. The sending component creates a messages, adds data to the message, and pushes the messages to a channel. The messaging system, or Message-Oriented Middleware (MOM)~\cite{curry2004:mom}, makes the message available to an appropriate component. The receiving component consumes the message and extracts the data from the message. 

\subsection{Enterprise Integration Patterns}
We use Enterprise Integration Patterns~\cite{hohpe2004enterprise}, or just \textit{patterns}, as a way to solve recurring architectural problems in the messaging context. The patterns are formulated in an abstract way to provide technology-independent design solutions. Besides, the patterns provide a common vocabulary to make technical discussion and reasoning easier to follow.

\section{Related Work}\label{sec:rw}
Several studies have focused on designing planning architectures, frameworks and service-oriented planning systems. Since these studies have appeared over a longer period (1997-2017), the range of architecture designs is also broad. A monolithic planning framework, called CPEF, is designed to offer several planning components~\cite{Myers_1999}. Due to the tight coupling of the components, the framework has a lack of flexibility. Usability, interoperability and reusability seem to be not considered. An implementation of the framework seem to have existed at the time of the publication.

PELEA, a generic architecture for planning, execution, monitoring, replanning and learning, is presented in~\cite{pelea2012}. The architecture seems to be a monolithic design with some modular structure where the main focus is on the processing of permanently incoming data. For usability, a dashboard is offered. Apart from an interface for the execution module, no other interfaces are described. An implementation of the architecture existed that focused on sensing and actuating.

A set of planning services and a service-oriented architecture for the domain of space mission are discussed in~\cite{Fratini2013}. The architecture does not focus on usability though some visualisation service are envisioned. To ensure interoperabilty, the standardisation of interfaces is discussed. The approach seems to offer a potential for rapid prototyping since many implementations already exist within the domain. The presented approach does not go beyond a theoretical discussion. 

The RuG planner is a system designed on the basis of an SOA and event-driven approach~\cite{kaldeli2016:planning}. Usability is not specifically investigated though it might have a potential for a high degree of user satisfaction due to the incorporating of industry needs. No particular interoperability considerations can be observed. The planner integrates an existing constraint-satisfaction solver, supporting reusability to some extent. Particular attention is paid to the scalability of the planner. The system is prototypically implemented and empirically evaluated to demonstrate its performance.

The SH planner is a system designed as a service-oriented architecture and offers a high degree of flexibility~\cite{70d0f3ca58114074a8723bd66eab370e}. Usability is enabled by a Web interface for modelling planning domains and problems. The use of RESTful interface offers a certain level of interoperability. No reusability is considered. The system is implemented as a prototype and used in the domains of ubiquitous computing~\cite{georgievski2017:pmc} and cloud deployment~\cite{georgievski2017:cloudapps}. 

Planning.Domains is a convenient platform solution that does not require any setup~\cite{Muise2016}. The solution is primarily intended for scientific and educational purposes. It provides several RESTful services that offer only technical interoperability. On the other hand, plugins are offered, allowing new functions to be integrated. The solution integrates an existing framework to solve planning problems. 

SOA-PE, a service-oriented architecture for sensing, planning, monitoring, execution, and actuation, is presented in~\cite{Feljan2017}. It draws architectural ideas from PELEA but transforms them into an SOA. The architecture seems to offer usability by providing appropriate interfaces. The architecture components are designed to communicate via some middleware for cyber-physical systems. Except for component distribution, no particular software design approaches or standards are observable. The architecture seems to have been prototypically implemented.

In contrast to these existing studies, our approach provides a planning architecture with a large collection of common planning capabilities. The architecture design is based on software design principles and patterns which have not been considered in the existing studies. Our prototype implementation represents a modern planning system that demonstrates the possibility of rapid prototyping and flexibility of composing planning capabilities. We also qualitatively assess our approach with respect to usability, interoperability and reusability. 

\section{Planning Capabilities}\label{sec:pc}
Before presenting the planning architecture, we first need to identify and collect important AI planning capabilities that will form the basis for the design of the architecture. In addition, we propose classifications of the identified AI planning capabilities to provide a systematic overview of the different kinds of properties, features, commonalities and differences among AI planning capabilities.

\subsection{Identification}
We want to identify and collect common AI planning capabilities necessary to design and implement a wide range of planning-based systems. To do this, we search existing literature describing planners, planning architectures and planning frameworks. We refer to these as \textit{planning artefacts}. In addition to those identified in related work, we select several prominent planners and systems, resulting in a total of 20 artefacts. We then employ content analysis and inductively identify planning functionality. The outcome is a collection of 18 AI planning capabilities. \cref{tab:pc:collection} shows an overview of the analysed planning artefacts and the capabilities identified per each artefact. Sometimes a planning capability service is described but not provided by the artefact itself, or it is not available in all versions of the artefact (indicated by \enquote{(X)}).

We determine three types of artefacts: planners, which are artefacts whose main capability is to solve planning problems (indicated by \enquote{P}); planning systems with an external planner, which are artefacts whose range of capabilities goes beyond the solving capability and integrate an external planner for the solving capability (indicated by \enquote{$\widehat{\mathrm{P}}-$}); and planning systems with a dedicated planner (indicated by \enquote{$\widehat{\mathrm{P}}+$}). 

A Converting capability performs transformation of planning data from one format to another without much complexity.
An Executing capability executes plan actions typically by using low-level commands or APIs.
A Graph-handling capability provides certain graph utilities that often needed or prove useful to planning techniques.
A Learning capability is commonly used for learning domain models or to aid the planning process.
Managing and orchestrating capabilities are used as routers or system handlers.
A Modeling capability represents an interface between users and planning systems. They are used to generate domains and problem instances.
Monitoring capabilities observe the world and execution of plan actions, and look for potential execution contingencies.
A Normalising capability is especially used in the context of heuristics. Also, some unifying steps can be added to this capability.
A Parsing capability is necessary to create programming-level objects from models specified in the planning languages.
Problem generating capabilities are used to automatically create planning problem instances from data representing the world, such as sensor data.
A Reacting capability enables handling unexpected events. Typically, this requires replanning.
Planners often use heuristics to find plans fast or of improved quality. The Relaxing capability supports the relaxation heuristics~\cite{Hoffmann2001}.
A Searching capability deals with search algorithms used to explore the space of planning states.
A Solving capability creates a plan and can utilise any number of supporting capabilities during the process.
In order to reuse domain specifications or other elements, Storing capabilities are required. 
A Verifying and Validating capability is designed to detect possible errors at an early stage.
When users or sensors are involved, it is necessary to filter incorrect inputs to reduce the system load. It is also used to validate the correctness of plans
A Visualising capability is a front-end functionality used to display charts, tables, and other statistics.

\begin{sidewaystable*}
  \centering
  \begin{tabular}{l|l|c*{17}{|c}|}
    \toprule
& & \multicolumn{18}{c|}{\textbf{Capability}} \\ \cmidrule{3-20}
\textbf{Name} & \textbf{Type} & 
\sgrROT{Parsing} &
\sgrROT{Modelling} &
\sgrROT{Solving} &
\sgrROT{Visualising} &
\sgrROT{Verifying \& Validating} &
\sgrROT{Executing} &
\sgrROT{Monitoring} &
\sgrROT{Managing} &
\sgrROT{Reacting} &
\sgrROT{Converting} &
\sgrROT{Searching} &
\sgrROT{Graph-Handling} &
\sgrROT{Relaxing} &
\sgrROT{Storing} &
\sgrROT{Learning} &
\sgrROT{Explaining} &
\sgrROT{Normalising} &
\sgrROT{Problem generating} \\ \midrule
CPEF  \cite{Myers_1999}
& $\widehat{\mathrm{P}}+$ & \sgrYes & \sgrYes & \sgrYes & \sgrYes & & \sgrYes & \sgrYes & & & & & & & & &
& 
& \sgrYes 
\\ \hline
FAPE  \cite{dvorak:hal-01138105}
& P & \sgrYes & \sgrYes & \sgrYes & \sgrYes & & \sgrYes & & & & & & & & & &
& 
& 
\\ \hline
Fast Downward  \cite{helmert-jair2006}
& P & \sgrYes & & \sgrYes & & & & & & & & \sgrYes & \sgrYes & \sgrYes & & &
& \sgrYes 
& 
\\ \hline
Fast Forward  \cite{Hoffmann2001}
& P & \sgrYes & & \sgrYes & & & & & & & & \sgrYes & \sgrYes & \sgrYes & & &
& \sgrYes 
& 
\\ \hline
GTOHP  \cite{Ramoul2017}
& P & \sgrYes &  & \sgrYes & & & & & & & & & & & & &
& 
& 
\\ \hline
Marvin  \cite{rintanen2001overview}
& P & \sgrYes & & \sgrYes & & & \sgrYes & & & & & \sgrYes & & \sgrYes & & &
& 
& 
\\ \hline
O-Plan2  \cite{Tate94o-plan2:an}
& $\widehat{\mathrm{P}}+$
& \sgrYes 
& \sgrYes 
& \sgrYes 
& \sgrYes 
& \sgrYes 
& \sgrYes 
& \sgrYes 
& \sgrYes 
& \sgrYes 
& 
& 
& 
& 
& 
& 
& 
& 
& \sgrYes 
\\ \hline
PANDA  \cite{bercher2014hybrid}
& $\widehat{\mathrm{P}}+$ & \sgrYes & \sgrYes & \sgrYes & \sgrYes & \sgrYes & \sgrYes & & \sgrYes & & \sgrYes & & \sgrYes& \sgrYes & & & \sgrYes
& \sgrYes 
& \sgrYes 
\\ \hline
PELEA  \cite{pelea2012}
& $\widehat{\mathrm{P}}+$ & \sgrYes & & \sgrYes & & & \sgrYes & \sgrYes & & & & & & & & \sgrYes &
& 
& \sgrYes 
\\ \hline
ROSPlanner  \cite{Cashmore2015}
& $\widehat{\mathrm{P}}-$ & \sgrYes & & (\sgrYes) & & \sgrYes & \sgrYes & & & & & & & & & &
& 
& 
\\ \hline
RuG Planner  \cite{kaldeli2016:planning}
& $\widehat{\mathrm{P}}+$ & \sgrYes & \sgrYes & \sgrYes & & & \sgrYes & \sgrYes & \sgrYes & \sgrYes & & & & & & &
& 
& \sgrYes 
\\ \hline
SH Planner  \cite{70d0f3ca58114074a8723bd66eab370e}
& $\widehat{\mathrm{P}}+$ & \sgrYes & \sgrYes & \sgrYes & & & \sgrYes & \sgrYes & \sgrYes & & \sgrYes & & & & \sgrYes & &
& \sgrYes 
& \sgrYes 
\\ \hline
SHOP  \cite{10.5555/1624312.1624357}
& P & \sgrYes & & \sgrYes & & & & & & & & & & & & &
& 
& 
\\ \hline
SHOP2  \cite{nau2003shop2}
& P & \sgrYes & & \sgrYes & (\sgrYes) & \sgrYes & & & & & & \sgrYes & \sgrYes & & & &
& 
& 
\\ \hline
SHOP3  \cite{Goldman2019}
& P & \sgrYes & & \sgrYes & & \sgrYes & & & & & & \sgrYes & \sgrYes & \sgrYes & & &
& \sgrYes 
& 
\\ \hline
\footnotesize{SIADEX}  \cite{fdez2006bringing}
& $\widehat{\mathrm{P}}+$
& \sgrYes 
& \sgrYes 
& \sgrYes 
& \sgrYes 
& 
& 
& \sgrYes 
& 
& \sgrYes 
& 
& 
& 
& 
& 
& 
& 
& 
& \sgrYes 
\\ \hline
SIPE-2  \cite{Wilkins2000}
& $\widehat{\mathrm{P}}+$
& \sgrYes 
& \sgrYes 
& \sgrYes 
& \sgrYes 
& \sgrYes 
& \sgrYes 
& \sgrYes 
& 
& 
& 
& \sgrYes 
& 
& 
& 
& 
& 
& 
& \sgrYes 
\\ \hline
SOA-PE  \cite{Feljan2017}
& $\widehat{\mathrm{P}}+$ & \sgrYes & & \sgrYes & & & \sgrYes & \sgrYes & \sgrYes & \sgrYes & & & & & & &
& 
& \sgrYes 
\\ \hline
SMP System  \cite{Fratini2013}
& $\widehat{\mathrm{P}}-$ & \sgrYes & \sgrYes & (\sgrYes) & \sgrYes & \sgrYes & \sgrYes & & & & & & & & \sgrYes & &
& 
& \sgrYes 
\\ \hline
UMCP  \cite{erol1994umcp}
& P
& \sgrYes 
& 
& \sgrYes 
& 
& 
& 
& 
& 
& 
& 
& 
& 
& 
& 
& 
& 
& 
& 
\\ \bottomrule
  \end{tabular}
  \caption[Capabilities Collection Overview]{An Overview of Planning Artifacts and Their Corresponding Capabilities. $\mathrm{P}$ indicates a planner, $\widehat{\mathrm{P}}-$ indicates a planning system with an external planner, and $\widehat{\mathrm{P}}+$ indicates a planning system with a dedicated planner. \enquote{X} indicates a support for a planning capability, and \enquote{(X)} indicates that a planning capability is not implemented or not available in all versions of the artifact.}
  \label{tab:pc:collection}
\end{sidewaystable*}


\subsection{Classification}
We classify the identified AI planning capabilities from two orthogonal views. The first view classifies the capabilities according to their operational competences, while the second view according to their technical properties.

\subsubsection{Operational View}\label{sec:pc:class:operational}
Our operational view for classification of AI planning capabilities is inspired by the following classes of capabilities defined in a business context for operational processes: core capabilities, enabling capabilities, and supplemental capabilities~\cite{leonard1995wellspring, prahalad1990g}. Core capabilities are defined as capabilities that provide access to a wide variety of domains, make a significant contribution to the perceived benefits, and are challenging to imitate~\cite{prahalad1990g}. Enabling capabilities are defined as capabilities that do not provide a competitive advantage on their own but are necessary for other capabilities~\cite{prahalad1990g}. Supplemental capabilities are easily accessible and provide a competitive advantage for the user. This class of capabilities often requires Enabling capabilities to run. 

By following these definitions, we classify planning capabilities as follows. The minimal set of Core planning capabilities consists of the Modeling, Solving and Executing capability. The set of Enabling planning capabilities consists of the Converting, Graph-handling, Managing, Monitoring, Normalising, Relaxing, Searching and Verifying \& Validating capabilities. The set of Supplemental planning capabilities consists of the Explaining, Learning, Parsing, Reacting, Storing and Visualising capabilities. 

\subsubsection{Technical View}\label{sec:pc:class:technical}
We now turn the view orthogonally to the technical perspective. It is known that patterns represent scientific and timeless approach to describe the behaviour and context of behaviour~\cite{alexander1979timeless}. The enterprise integration patterns instantiate such an approach~\cite{hohpe2004enterprise}, which we apply on the AI planning capabilities to classify them into appropriate classes. 

Before we map AI planning capabilities to patterns, we want to reduce the complexity of the planning process and its relevant aspects. To achieve this, we view the process as a black-box pipeline based on the Pipes and Filters pattern~\cite{hohpe2004enterprise}. The main components of this pattern are filter nodes, source nodes, and sink nodes. Filter nodes are processing units of the pipeline. Source nodes provide input data to the pipeline, while sink nodes collect the outcome from the end of the pipeline. By using these types of nodes and analysing the AI planning capabilities, we observe that each capability can be reduced to one of the node types. The left-hand side of \Cref{tab:pc:technical:patterns} gives an overview of the reduction of the planning capabilities to node types.

\begin{table*}[t]
  \centering
  \begin{tabular}{l|c|c|c||l|l}
    \toprule
\textbf{Capability}  & \multicolumn{3}{c||}{\textbf{Node Type}} & \textbf{Pattern} & \textbf{Class} \\
 & Source & Filter & Sink &  & \\ \midrule
  Converting 	    & & \sgrYes &          & \textit{Message Translator} 	    & Transforming \\ \hline
  Executing 	    & & & \sgrYes          & \textit{Service Activator} 	    & Endpoint \\ \hline
  Explaining 	    & & & \sgrYes          & \textit{Message Endpoint} 		& Endpoint \\ \hline
  Graph-Handling   & & \sgrYes &          & \textit{Message Translator}	    & Transforming \\ \hline
  Learning 	    & \sgrYes & \sgrYes &  & \textit{Message Endpoint} 		& Endpoint \\ \hline
  Managing 	    & & \sgrYes &          & \textit{Process Manager} 		& Routing \\ \hline
  Modelling 	    & \sgrYes & &          & \textit{Message Endpoint} 		& Endpoint \\ \hline
  Monitoring 	    & & \sgrYes &          & \makecell[l]{\textit{Control Bus},\\ \textit{Wire Tap}} 	& System-Management \\ \hline
  Normalising 	& & \sgrYes &          & \textit{Normalizer} 			    & Transforming \\ \hline
  Parsing 		    & & \sgrYes &          & \textit{Message Translator} 	    & Transforming \\ \hline
  Problem generating    & & \sgrYes &          & \textit{Content Enricher} 	    & Transforming \\ \hline
  Reacting 	    & & \sgrYes &          & \textit{Event-Driven Consumer}   & Endpoint \\ \hline
  Relaxing 	    & & \sgrYes &          & \textit{Message Translator} 	    & Transforming \\ \hline
  Searching 	    & & \sgrYes &          & \textit{Content Filter} 		    & Transforming \\ \hline
  Solving 		& & \sgrYes &          & \textit{Message Translator} 	    & Transforming \\ \hline
  Storing 		& & & \sgrYes          & \textit{Message Store} 		    & System-Management \\ \hline
  Verifying \& Validating    & \sgrYes & \sgrYes &  & \makecell[l]{\textit{Content Based Router},\\ \textit{Detour}} & Router \\ \hline
  Visualising     & \sgrYes & & \sgrYes  & \makecell[l]{\textit{Polling Consumer},\\ \textit{Event-Driven Consumer}}	& Endpoint
\\ \bottomrule
  \end{tabular}
  \caption{Technical Classification of AI Planning Capabilities}
  \label{tab:pc:technical:patterns}
\end{table*}

With this knowledge, we map the AI planning capabilities to appropriate patterns. The right-hand side of \Cref{tab:pc:technical:patterns} gives an overview of the mapping. The resulting patterns can be grouped into four classes, namely Endpoint, Transforming, System Management and Routing. The Endpoint class includes patterns that enable software components connect to a messaging system for the purpose of sending and receiving messages. The Transforming class encompasses patterns that translate, enrich, normalise or perform any other form of transformation of messages. The System Management class contains patterns that monitor the flow and processing of messages and deal with exceptions and performance bottlenecks of the underlying system. The Routing class includes patterns that enable decoupling a component that sends a message from one that receives the message. As a consequence of this pattern grouping, AI planning capabilities can be also categorised in the four classes, providing the technical view of classification, as shown in \Cref{tab:pc:technical:patterns}.

\begin{figure}
  \centering
  \includegraphics[width=\linewidth]{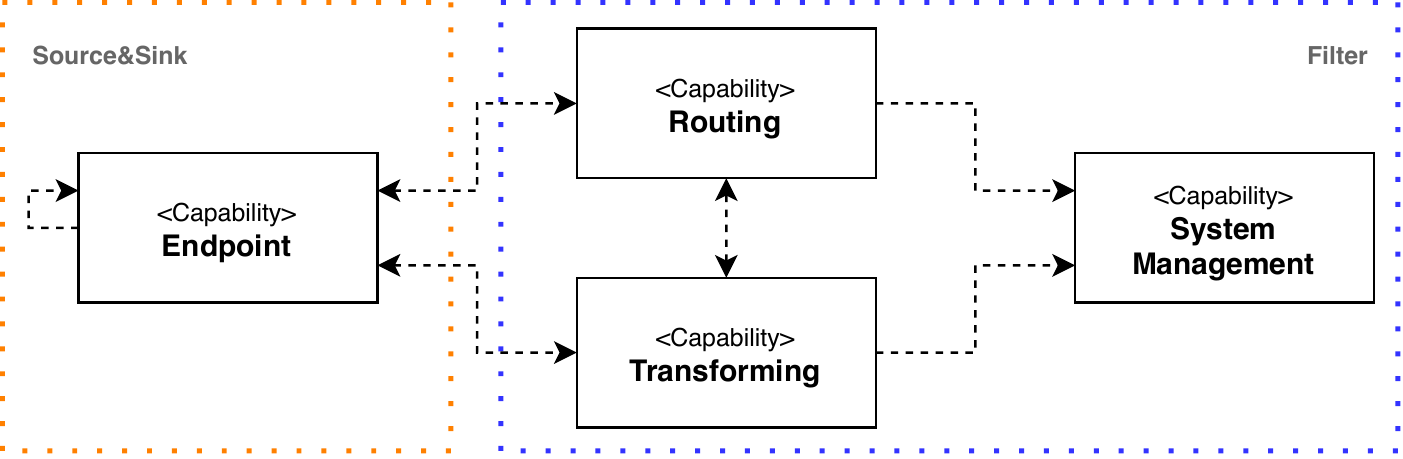}
  \caption{Relationships Between Pattern Classes and Node Types}
  \label{fig:pc:evaluation:technical:classification}
\end{figure}

The connections among the pattern classes and their relationship with the node types is shown in \cref{fig:pc:evaluation:technical:classification}. Endpoints represent either source nodes or sink nodes. Such endpoints can call each other or call filter nodes from the Routing or Transforming class.
System Management also monitors routing and transforming capabilities.

\section{Architecture Design}\label{sec:design}
With the collection of planning capabilities and the classifications, we can design a fully operational planning architecture. Next, we determine the requirements of interest and structure of the planning architecture, and then we provide some general and more specific consideration for the architecture design.

\subsection{Capability Requirements}\label{sec:ad:c_req}
While current planning artefacts provide powerful means to search for and execute plans, they lack qualities as software systems. Planning artefacts are often functionally overloaded to meet the goals, which increases the complexity and reduces the level of usability, which is the first requirement of our interest. 

Since real-world planning problems are too complex to be solved by a solving capability only, the problems are often decomposed into sub-problems that need to be handled by separate planning capabilities, such as modelling, execution and reaction. These capabilities in turn must communicate among each other and with external services, thus exchange messages. So, the second requirement of interest is interoperablity.

Many planning artefacts require the same planning capabilities for full functionality. For instance, numerous planning artefacts require a PDDL parsing capability. Currently, the planning artefacts either implement their own version of a parsing capability or integrate full planners to use the planners' parsing functionality. So, enabling reusability of planning capabilities is the third requirement we consider in the design of the planning architecture. 

\subsubsection{Usability}
Usability starts on the interaction side. Independent of the user's skills, the user should be able to use the planning capabilities. Thus, a certain level of \enquote{ease of use} is mandatory for planning capabilities to be usable. With good usability, it does not matter if a planning capability is a front-end or back-end service; the appropriate usability criteria are the same. According to the ISO~9241-11 standard~\cite{ISO-9241-11-2018}, we need to maximise the effectiveness, efficiency and user satisfaction. 

\subsubsection{Interoperability}
Interoperability is majorly achieved by using unified models on every endpoint. So, the concepts for loose coupling and messaging are mandatory. According to~\cite{Delgado2013}, four levels of interoperability should be considered for the design: organisational, semantic, syntactic and connectivity level. On the organisational level, we must ensure that each planning request follows a purpose. This level contains a strategy, choice of interacting component, and outsourcing parts of the functionality to other services. On the semantic level of interoperability, we have to ensure that both interacting services interpret a message or request in the same way. This means a domain of values, relevant rules, and/or choreography must be provided to ensure equivalent handling between interacting components. On the syntactic level, we need to focus on the notation (format) of a planning request. The message must follow the structure according to defined schematics. In most cases, RESTful Application Programming Interfaces (APIs) or messaging channels use JSON or XML as a message format. On the connectivity level, we need to deal with a protocol as an essential step for transferring a planning request. This level requires a selection of a message protocol and routing to the correct endpoint. Note that, to achieve the best possible interoperability, planning capabilities have to aim for the best interoperability on each of the levels.

\subsubsection{Reusability}
To mitigate or reduce the development effort for new AI planning capabilities, the reusability of existing planning concepts and services is indispensable. To allow reuse of planning capabilities, the number of dependencies has to be minimal. This requirement is also related to the loose coupling autonomies (see Section~\ref{sec:bg:soc-soa}). To support the reusability of planning capabilities, we consider the following quality characteristics: portability, flexibility, and understandability. For details on these characteristics of reusability, we refer to~\cite{Cardino1997}.

\subsection{Decentralised Approach}
Our choice for the design of the planning architecture falls to a decentralised approach as it offers the best properties to achieve loose coupling. First, the communication between capabilities based on messaging in the publish-subscribe stack is most often characterised by good performance, which is especially useful in systems with a high load and several clients, such as in energy smart systems~\cite{georgievski2017:pmc}. Second, decentralised planning systems may scale very well horizontally since each planning capability would have its own incoming and outgoing topic. Furthermore, the automatic configuration detection in MOM helps to add new planning capabilities during a system's runtime. The most crucial point is that planning capabilities themselves can request other capabilities directly without another gateway, checking the availability first. Finally, since it is essential to increase reliability and not violate the principles of services, the planning capabilities must be designed and implemented to be stateless. Therefore, our choice for the design of the planning architecture falls to the decentralised approach.

\subsection{Planning Architecture}\label{sec:design:architecture}
We design the planning architecture considering general service requirements (see~\cite{1407782}). The architecture aims at defining a composition of planning capabilities and thereby facilitating their implementation as services. Our proposal for the design of the planning architecture is based on the \textit{Hub-and-Spoke} pattern~\cite{hohpe2004enterprise}. Usually, this kind of architecture pattern is not performing well when it comes to scaling. However, since MOM provides the hub, selecting the best-fitting technology for MOM can help overcome this issue. The usage of this design pattern also provides an easy solution to connect new planning capabilities with each other using the hub.

\begin{figure}
  \centering
  \includegraphics[width=\linewidth]{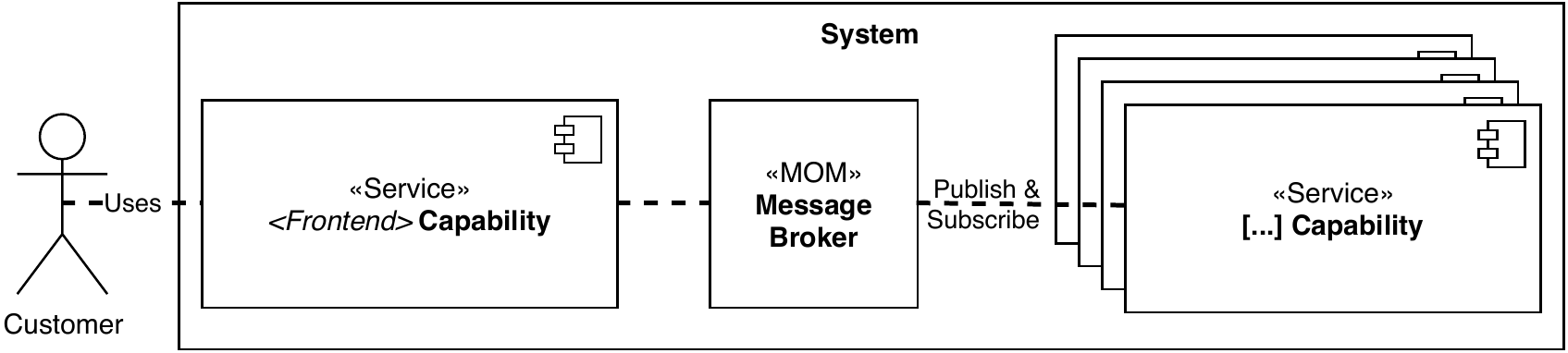}
  \caption{Abstract Overview of the Planning Architecture}
  \label{fig:design:abstract_architecture_messaging}
\end{figure}

\cref{fig:design:abstract_architecture_messaging} shows an abstract overview of the planning architecture. The architecture consists of a front-end service, MOM, and various back-end services. Due to the ease of implementation, we propose communication of the front-end capabilities based on RESTful Web services for synchronous calls and Web Sockets for asynchronous messaging. Using these standard Web communication concepts also enables the simple integration of security features (e.g., HTTPS and JWT on the header). 

\begin{figure*}
  \centering
  \includegraphics[width=.8\linewidth]{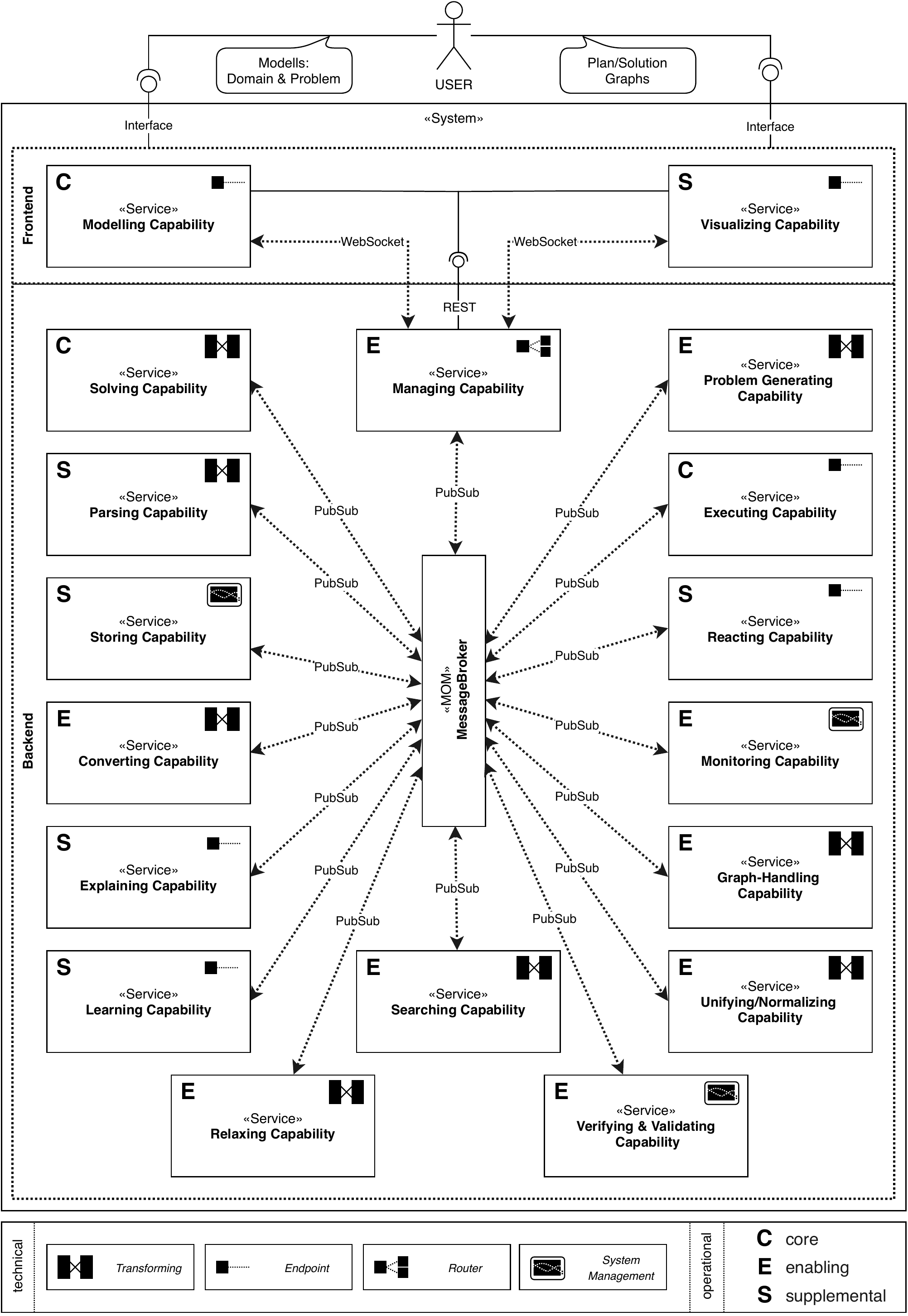}
  \caption{Overview of the Planning Architecture}
  \label{fig:design:architecture_messaging}
\end{figure*}

\cref{fig:design:architecture_messaging} shows the complete architectural design. It can be observed that the architecture provides great opportunities since the operational and technical classes also specify the context of use. The range of functionality can be extended significantly by implementing different capability service instances for the same capability class. Some capabilities require different interfaces to work in a sufficient way. For instance, the Executing capability requires interfaces for action invocations. Besides this, the Storing capability requires a database to do not violate the stateless principles of services. Since front-end capabilities require a bridging service to access the MOM data, a Managing capability is utilised. All other capabilities are connected using messaging on a publish-subscribe basis.

User interaction is also considered by accounting for appropriate graphical user interfaces through which information can be passed to the system. This information mainly refers to models of domains and problems, or configurations. Additionally, a Visualisation capability is envisioned to display relevant results (e.g., sequential plans) and to provide monitoring figures or statistics. The design does not put any restrictions on the use of third-party API.

\subsubsection{Messaging}
We propose the following template for naming MOM topics:
\begin{center}
    {\footnotesize\verb+<ver>.<capability-class>.<capability-name>#<name>+}
\end{center}

\noindent The last part of the template ({\footnotesize\verb+#<name>+}) defines the handling of instances of a capability. Note that a capability instance refers to an implementation instance rather than a service instance for high availability. This value should not be a unique instance ID since horizontal scalability is necessary. In case of multiple instances of the same planning service, a common name should be used to enable parallel execution of tasks. Therefore, the architecture does not define any implementation detail. Only the existence of an additional sub-topic or flag is important.

To avoid blocking, outgoing topics are not provided since this would encourage polling endpoints. It is necessary to specify a response topic in a request message to route the message after its processing dynamically. The underlying pattern is \textit{Request-Reply} with an extension of the \textit{Return Address} pattern~\cite{hohpe2004enterprise}.

\subsubsection{Statelessness}
If a capability service needs to assign a sub-task of its planning task to another capability, the state of the process must be stored temporarily so that the capability service becomes stateless and nonblocking after the assignment (i.e., sending a message). To maintain the IDEAL properties of the future system (see~\cite{Fehling2014, Breitenbucher2019}), the use of an isolated state is necessary. Therefore, the planning capabilities that call other capabilities to place requests must store the state externally. To merge the results of sub-tasks in the corresponding capability service, a so-called \textit{correlation identifier} is used to guarantee the correctness of the correlation~\cite{hohpe2004enterprise}.

\subsubsection{Handling of Requests}
We now describe the adopted integration patterns per technical class. These patterns can be applied to all capability services of the respective class.

\subsubsection*{Endpoint planning capabilities}
Since Endpoint planning capabilities can occur in different variants, a general pattern is not applicable. Therefore, a distinction between front-end and back-end services is necessary.

\begin{figure}
  \centering
  \includegraphics[width=\linewidth]{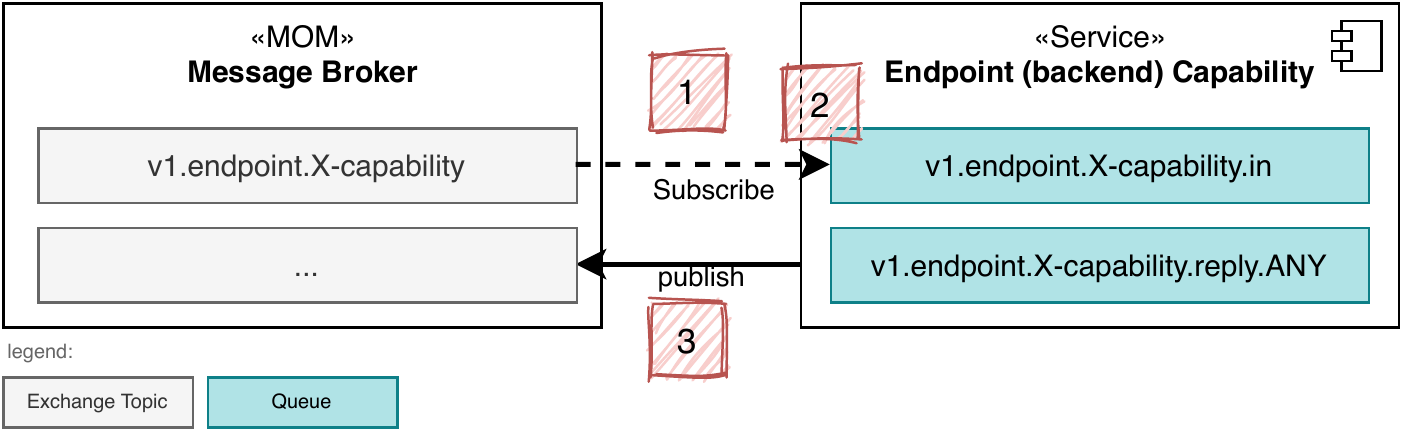}
  \caption{Back-End Endpoint Capability}
  \label{fig:design:capability:endpoint:backend}
\end{figure}

\paragraph{Back-end capabilities}
All planning capabilities within the Endpoint class that do not have to process direct user input can be considered back-end capabilities. \cref{fig:design:capability:endpoint:backend} shows an example for the processing of a planning request. The following steps are performed (also highlighted in \cref{fig:design:capability:endpoint:backend}):

\begin{enumerate}
	\item The back-end service has an established subscription to its incoming exchange topic and starts processing the message.
	\item The request is received and placed in a queue. All the requests will be processed after each other (i.e., a first-in-first-out order). The results are pushed to the corresponding reply queue.
	\item The response is taken from the reply queue and send to the corresponding topic.
\end{enumerate}

\begin{figure*}
  \centering
  \includegraphics[width=.8\linewidth]{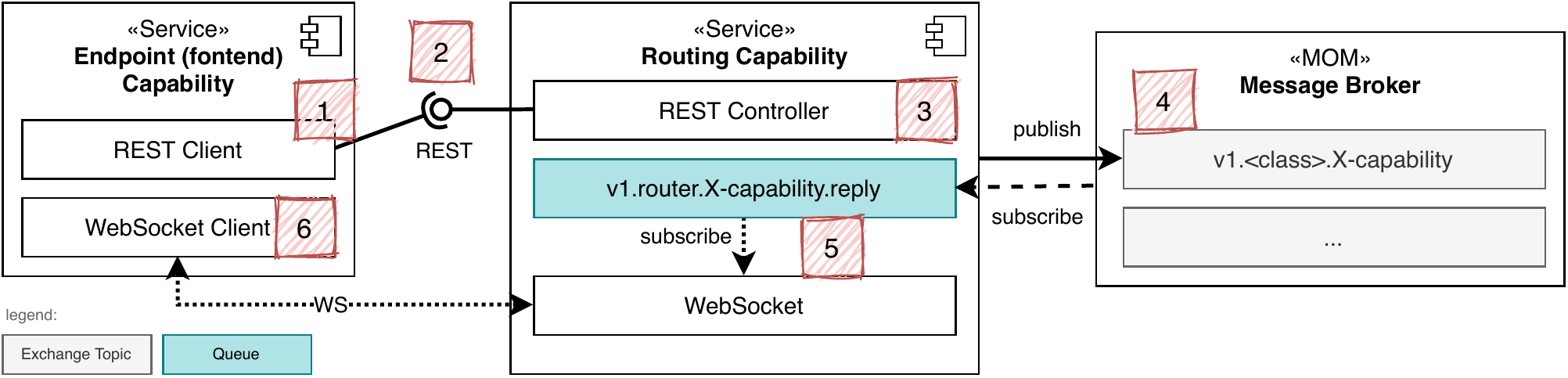}
  \caption{Front-End Endpoint Capability}
  \label{fig:design:capability:endpoint:frontend}
\end{figure*}

\paragraph{Front-End capabilities}
All planning capabilities within the Endpoint class that have to process direct user input can be considered front-end capabilities. \cref{fig:design:capability:endpoint:frontend} shows an example for the processing of a user request. The following steps are performed (also highlighted in \cref{fig:design:capability:endpoint:frontend}):

\begin{enumerate}
	\item The front-end capability service receives new information from user input. All information is placed at the request body of a RESTful POST message.
	\item The front-end capability service calls the RESTful endpoint of a routing capability.
	\item The Routing capability service process the request and sends it to a Solving capability using MOM.
	\item After the system processed the request, the response or error is passed back to the routing capability. A Web Socket is used to send the asynchronous response back to the front-end.
	\item The front-end receives the response or a custom-error and processes it (e.g., visualises it).
\end{enumerate}

\subsubsection*{Transforming capabilities}
\begin{figure*}
  \centering
  \includegraphics[width=.8\linewidth]{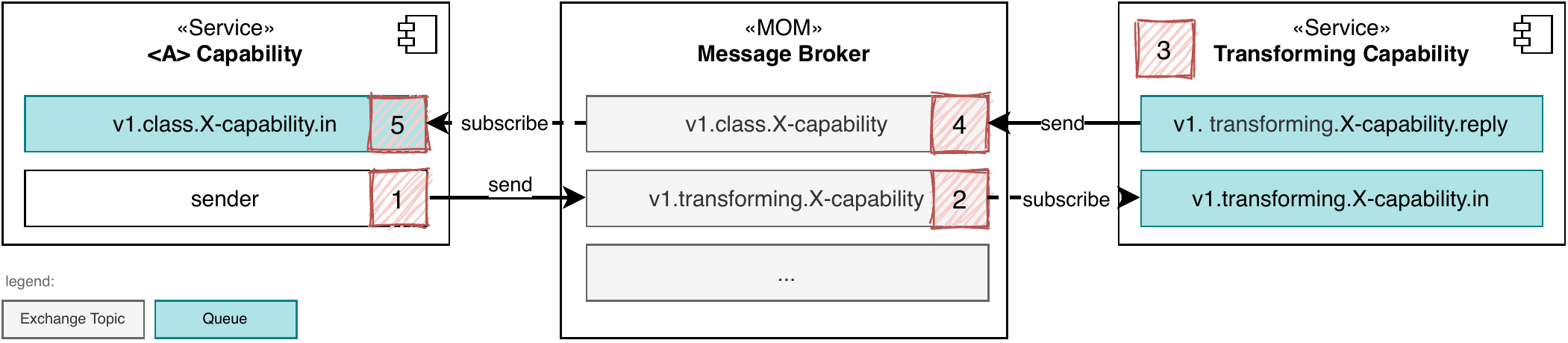}
  \caption{Transforming Capability}
  \label{fig:design:capability:transforming}
\end{figure*}

Since all planning capabilities in the Transforming class are identical from an architectural point of view, one pattern can be applied to all of them. \cref{fig:design:capability:transforming} shows a processing of a simple request by a transforming capability. The following steps are performed (also highlighted in \cref{fig:design:capability:transforming}):

\begin{enumerate}
  \item Service A creates a new request and pushes the message to the corresponding transforming topic in MOM.
  \item The transforming service has an established subscription to its incoming topic, and the message is pushed to the queue of the transforming capability.
  \item The transforming service process the request and sends it to the given response topic. 
  \item The response is placed on service A's capability topic in  MOM.
  \item Service A has an established subscription to its incoming topic and starts processing the message.
\end{enumerate}

Since the architecture allows dynamic routing, the transforming capability is free to decide if any other transformation is necessary to reach the requested response format.

\subsubsection*{System Management Capabilities}
System management and monitoring capabilities differ slightly at the architecture level. Only the origin of the information varies within MOM. An example of behavior of the system management capability includes a step in which the service establishes a subscription to at least one topic and another step to process a message.

\subsubsection*{Routing capabilities}
\begin{figure*}
  \centering
  \includegraphics[width=.8\linewidth]{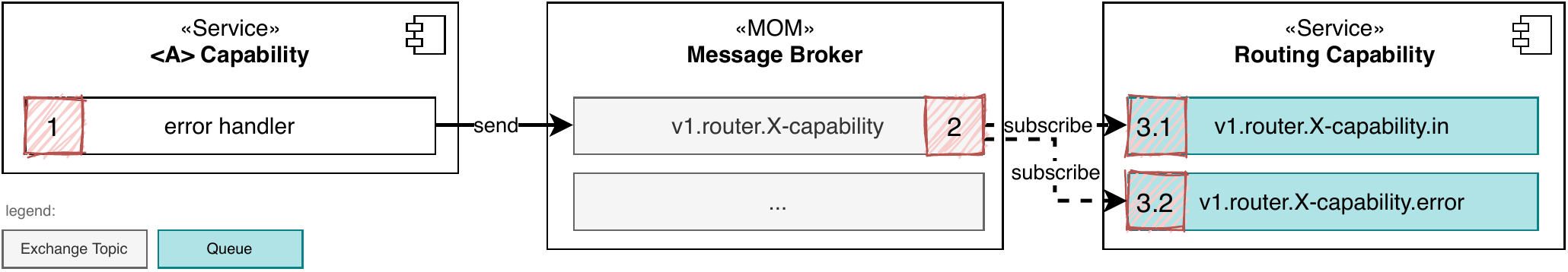}
  \caption{Routing Capability}
  \label{fig:design:capability:router}
\end{figure*}

Since there is only one capability classified as a routing capability yet, one architecture pattern can be applied. The basic functionality is similar to the one for the front-end endpoint capability. Here we focus on the global error handling that is done by the Routing capability. \cref{fig:design:capability:router} shows the architecture pattern and behavior for the Routing capability. The following steps are performed (also highlighted in \cref{fig:design:capability:router}):

\begin{enumerate}
	\item Service A throws a custom error on processing a request. The request is handled and sent to the routing capability topic.
	\item The routing service consumes the message from the topic and pushes it to the correct queue.
	\item Normally the request is pushed to the in-queue (step $3.1$). In this example, the message is an error, so the error-queue is responsible (step $3.2$). The response or error is processed depending on the specific implementation.
\end{enumerate}

\section{Implementation}\label{sec:impl}
\begin{figure*}[!t]
  \centering
  \includegraphics[width=.8\linewidth]{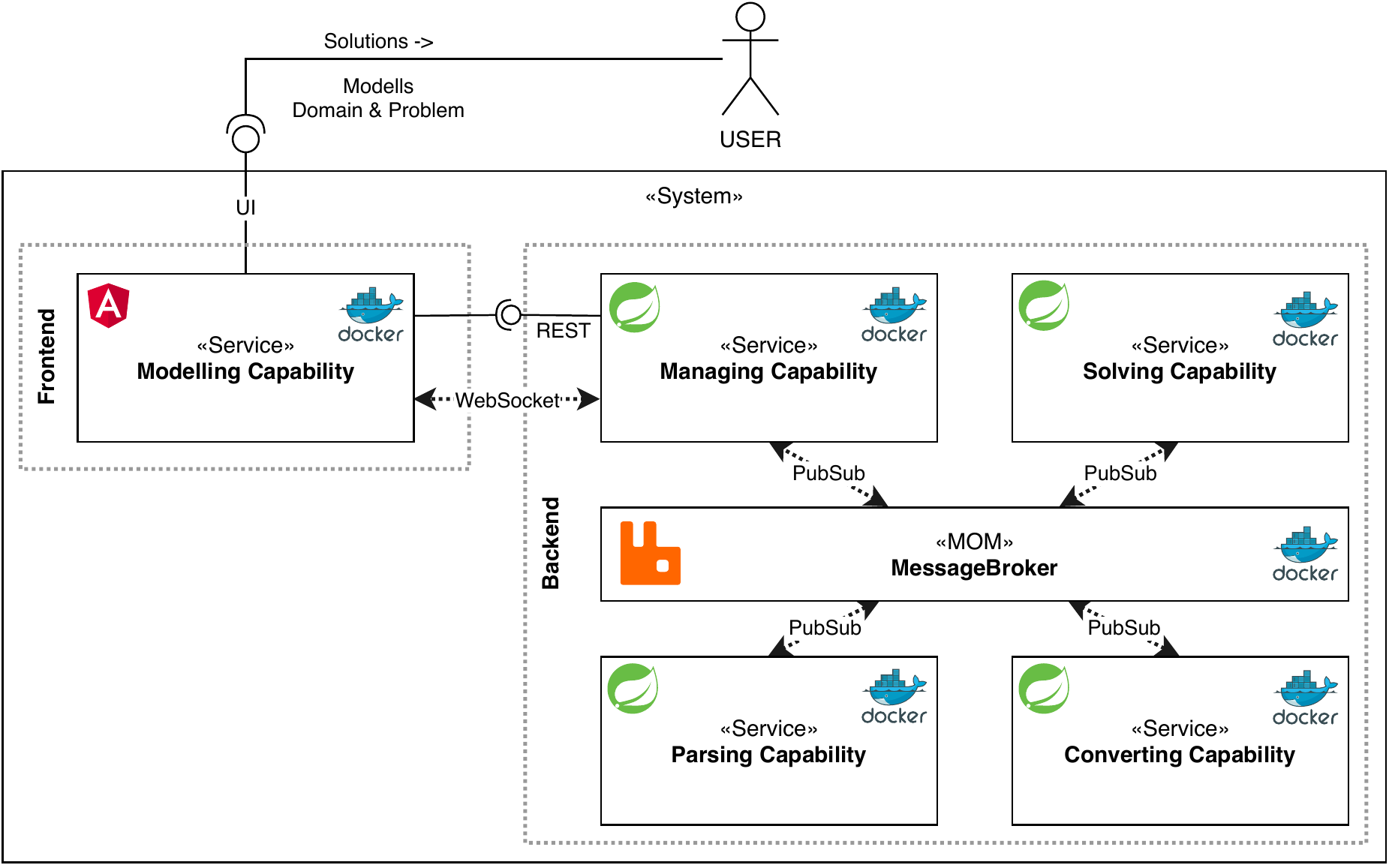}
  \caption{Deployment Diagram of the Prototype Planning System}
  \label{fig:prototype:overview}
\end{figure*}

As a proof of concept, we implemented a prototype planning system whose deployment diagram is shown in \cref{fig:prototype:overview}. It consists of five planning services that provide a fully functioning planning system. Besides the general division into front-end and back-end services, the technologies used for the realisation of each service are also shown. We implemented the modelling capability using Angular\footnote{\url{https://angular.io}} and Typescript.\footnote{\url{https://www.typescriptlang.org}} For the back-end capabilities, we used Spring Boot\footnote{\url{https://spring.io/projects/spring-boot}} and Kotlin.\footnote{\url{https://kotlinlang.org}} As message-oriented middleware, we use RabbitMQ.\footnote{\url{https://www.rabbitmq.com}}

We used reactive programming, among other things, to be able to handle asynchronous processes in a non-blocking manner. Special attention is paid to dynamic routing to avoid the classical ping-pong process and to increase reusability. We configured RabbitMQ to give each planning service its message queue for incoming tasks. As RabbitMQ provides built-in filtering of messages in a topic (the so-called routing key), we use this feature to not compromise the extensibility of the same capability classes (e.g., HDDL parsing capability). Each queue can be configured to accept only the tasks with the appropriate routing key from the topic. This enables creation of new planning services and their integration into the running system. Each service is able to create related topics and register new queues in the messaging system.

Each planning service can be delivered as a Docker\footnote{\url{https://www.docker.com}} image with a single command. The entire planning system can be deployed using Docker-Compose. By utilising runtime environment parameters, a predefined profile can be selected to allow the deployment of the system to different stages.

To use a minimal amount of process-relevant information in services, we use the \textit{Routing Slip} pattern. This pattern allows dynamic routing of messages across multiple services. Implementing the process within a capability would cause reusability to suffer. The pattern is implemented using a call stack. After initially receiving a system request, the system's expected endpoint (topic name and expected type) is first pushed followd by the initial service pushed by topic name and schema (or only a schema ID if a type registry exists). Typically, the Solving service is the initial capability. In it, a process is pushed into the stack in the form of sequential steps. Later, the invoked capabilities can extend the process arbitrarily by adding more process steps in the stack. Each capability removes its entry from the stack and sends the result to the next entry above in the stack.

In the following, we provide insights into the implementation of each prototype capability.

\subsection{Managing Capability}
We use the Managing capability to integrate the front-end and back-end. The capability is responsible for handling the user requests and delivering results to users. The Managing service also implements basic error-handling that uses the correlation identifier of each request. Due to the decentralisation of the message flow, we set up an error queue that transmits the corresponding status to the user when a certain planning capability step fails. Finally, the Management capability store interfaces of the available capabilities, which can be useful to other planning capabilities (see Solving capability). Currently, only an in-memory solution is integrated, which cannot be used by other capabilities. Note that a list of all available capabilities is already realised in the prototype by using the RabbitMQ API.

For communication with the Modelling capability, the Managing capability provides a RESTful interface. For connection to RabbitMQ, a corresponding exchange topic is created. To efficiently handle asynchronous system responses (i.e., results and errors), the Managing capability implements a Web Socket. Non-blocking asynchronous event streams are processed using Reactor.\footnote{\url{https://spring.io/reactive}}

\subsection{Parsing Capability}
The Parsing capability converts planning problems specified in some syntax into programming-level objects. This service requires implementation of a \verb|ParsingRequest| interface, which includes syntactic models of a planning domain and problem instance. To reduce the size of messages, the input should be given as \verb|base64| strings. The input is interpreted according to the stored information in the call stack. To manage multiple syntactic forms of inputs to this service (e.g., PDDL, HDDL, SHOP), we use the \textit{Strategy} pattern, which allows encapsulating a set of interchangeable algorithms individually~\cite{lavieri2019:java-patterns}. The basic approach towards the implementation of this service consists of converting internal data models into wrapped data models which are in turn serialised and provided as JSON messages. The conversion is accomplished by the Converting capability.

The service implementation parses PDDL planning problems. This is accomplished using the PDDL4J library~\cite{Pellier2018}. For other alternatives, see Section~\ref{sec:discussion}.

\subsection{Converting Capability}
We use the converting capability to demonstrate the dynamic routing and independent insertion of intermediate steps, such as encoding of planning problems. The encoding step transforms a given planning problem into its final form. Such a transformation is typical for most PDDL-based planners~\cite{Pellier2018}.

The Converting service requires implementation of the \verb|EncodingRequest| interface. Since our prototype completely supports planning problems specified in PDDL, we implemented the encoding of PDDL to a finite domain representation using the PDDL4J library. We created a \verb|PddlEncodedProblem| class that follows the object structure of the library. We wrap the whole data structure by custom classes to avoid serialisation problems. This implementation choice preserves the representation of a given planning problem as a causal domain transition graph, which provides an efficient way to determine which propositions are accessible from the current state~\cite{Pellier2018}.

\subsection{Solving Capability}
The Solving capability requires implementation of a \verb|SolvingRequest| interface with four arguments about the chosen planner, chosen planning language, domain, and problem instance. A user can select a planner and planning language. The service subsequently decides on its own authority which planning capabilities are to be used. That is, the \verb|solveProblem()| method is called, which sends a request to the Parsing capability. This connection is currently preconfigured, but the ultimate objective is to search a repository for required interfaces and select an appropriate service interface. The cornerstone for this is already implemented in the Managing capability. 

After receiving a reply from the Converting capability, the solving of the given planning problem begins. Since the data model may not contain a planner field for the response, the call-stack state is used at this point. Depending on the embedded planner, different types of plans can be produced (e.g., sequential plans or partially ordered plans). All results are forwarded according to the call stack.

In our case, the user can select planners with different search strategies and heuristics as implemented in the PDDL4J library. For other alternatives, see Section~\ref{sec:discussion}.

\subsection{Modelling Capability}
The Modelling capability provides a Web interface that enables a user to model planning problems, select a planner, send a solving request, and inspect a found plan. It also provides login, monitoring and administration functionality, such as overview of the system and states of capabilities. In case of errors, the Modelling capability provides the stack trace and a reference to the corresponding service. The modelling of planning problems is enabled by a WEB IDE implemented using ACE\footnote{https://ace.c9.io/} with syntax highlighting and folding provided by the WEB planner~\cite{magnaguagno2017:webplanner}.

\section{Qualitative Assessment}\label{sec:benefits}
We now provide a qualitative assessment of our approach. We analyse the usability, interoperability, and reusability of the planning architecture/prototype in comparison with a typical AI planning artefact. To visualise the comparisons of the quality data, we use radar charts. Each quality attribute is divided into several indicators each of which gets a score assigned. The score describes the general suitability of the architecture/artefact with respect to the indicator. We use $\{ -, O, +\}$ as a score range, where each score is correlated with a value as follows: \enquote{$-$}$ \equiv 0.0$; \enquote{$O$}$ \equiv 0.5$; and \enquote{$+$}$ \equiv 1.0$.

\subsection{Usability}
\begin{figure}
  \centering
  \includegraphics[width=\linewidth]{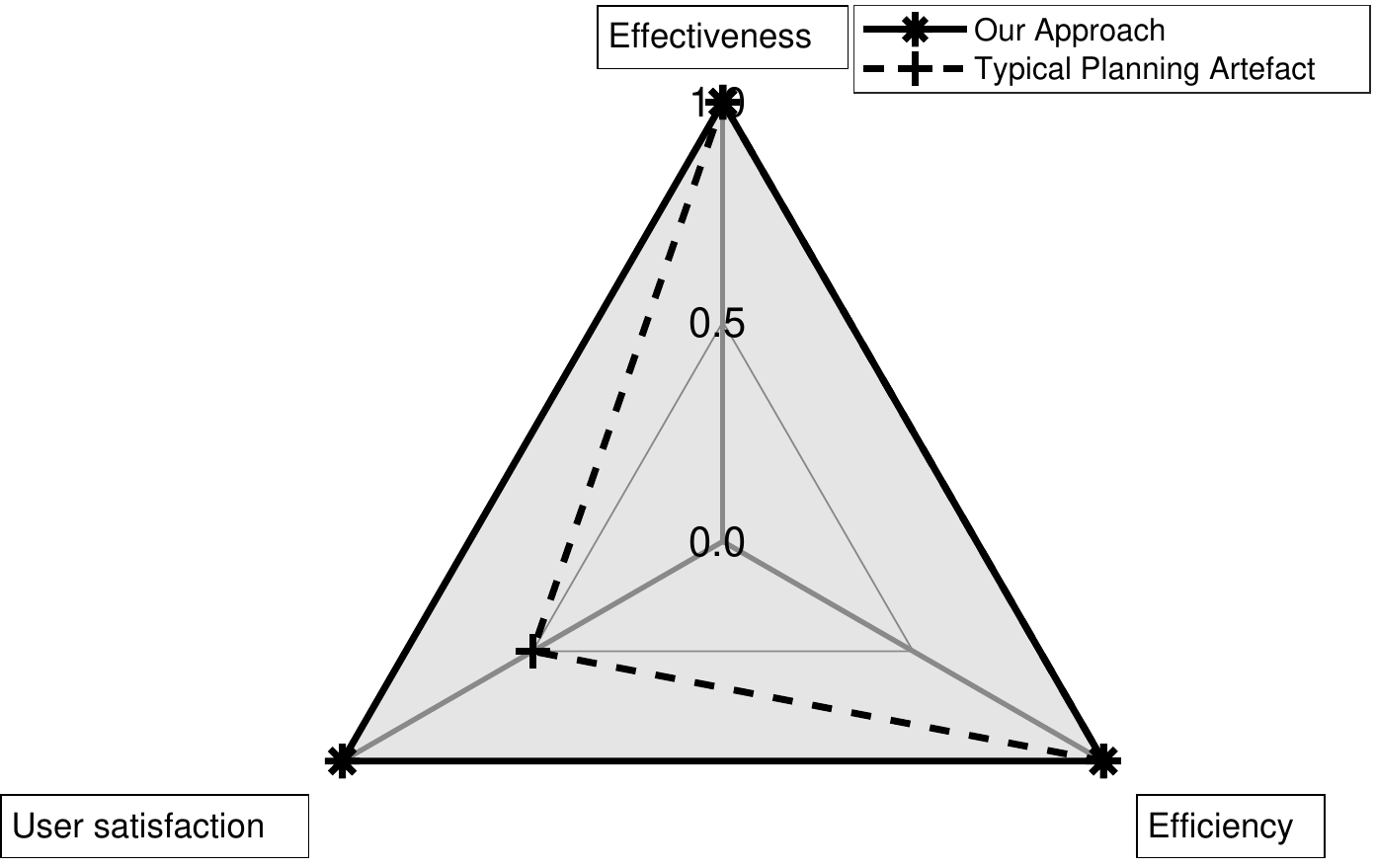}
  \caption{Radar Chart of Usability Indicators}
  \label{fig:evaluation:usability:plot}
\end{figure}

Since we are not completely interested in user interfaces, we focus on usability from a developer's point of view. We assess usability considering three indicators: effectiveness, efficiency and user satisfaction. \cref{fig:evaluation:usability:plot} shows the scores of these usability indicators. It can be observed that our approach scores equally or better than a typical planning artefact. We discuss each usability indicator next.

\paragraph{Effectiveness}
We see the effectiveness as a composition of task effectiveness, task completion and error frequency~ \cite{Abran2010}. Since we only require existing planning artefacts to be distributed over corresponding planning capabilities, the operation and behaviour of such artefacts would not be affected. As a consequence, the task effectiveness and task completion of integrated planning artefacts are not compromised. If our planning services are distributed over a network, the probability to have a high error frequency may grow due to service unavailability. This would not be the case with typical planning artefacts. However, our planning architecture and prototype support resilience by allowing multiple service instances to run and using a messaging system. We can conclude that the general effectiveness of our approach is as high as a typical planning artefact without network dependencies. Thus, both score \enquote{$+$}.

\paragraph{Efficiency}
We use two indicators for the efficiency. The first one is related to the first startup of a system. Since most planning artefacts have a large number of dependencies, the first startup is typically time-consuming. On the other hand, our approach is a platform solution that only requires a Web browser available in any standard computing device.

The second indicator is related to system runtime from the moment of submission of a planning request. Under the assumption of identical computing power, our approach might be slower that a typical planning artefact due to serialisation and messaging, which may require additional computing power. We should also consider that our architecture design offers the possibility of parallelisation. As a consequence, our approach and a typical planning artefact both score \enquote{$+$}.

\paragraph{User Satisfaction}
We can assume a high degree of user satisfaction can be achieved by providing front-end capabilities.
However, due to the ability to extend capabilities, our architecture offers a significant advantage over a typical planning artefact. So, the probability of being able to deliver a desired function is correspondingly higher than with a typical planner. Since most existing planners only provide or require a command-line interface, the required knowledge and user involvement is high, and errors due to wrong interaction can occur more easily. Therefore, our approach scores \enquote{$+$}, while a typical planning artefact scores \enquote{$O$} for user satisfaction. 

\subsection{Interoperability}
\begin{figure}
  \centering
  \includegraphics[width=\linewidth]{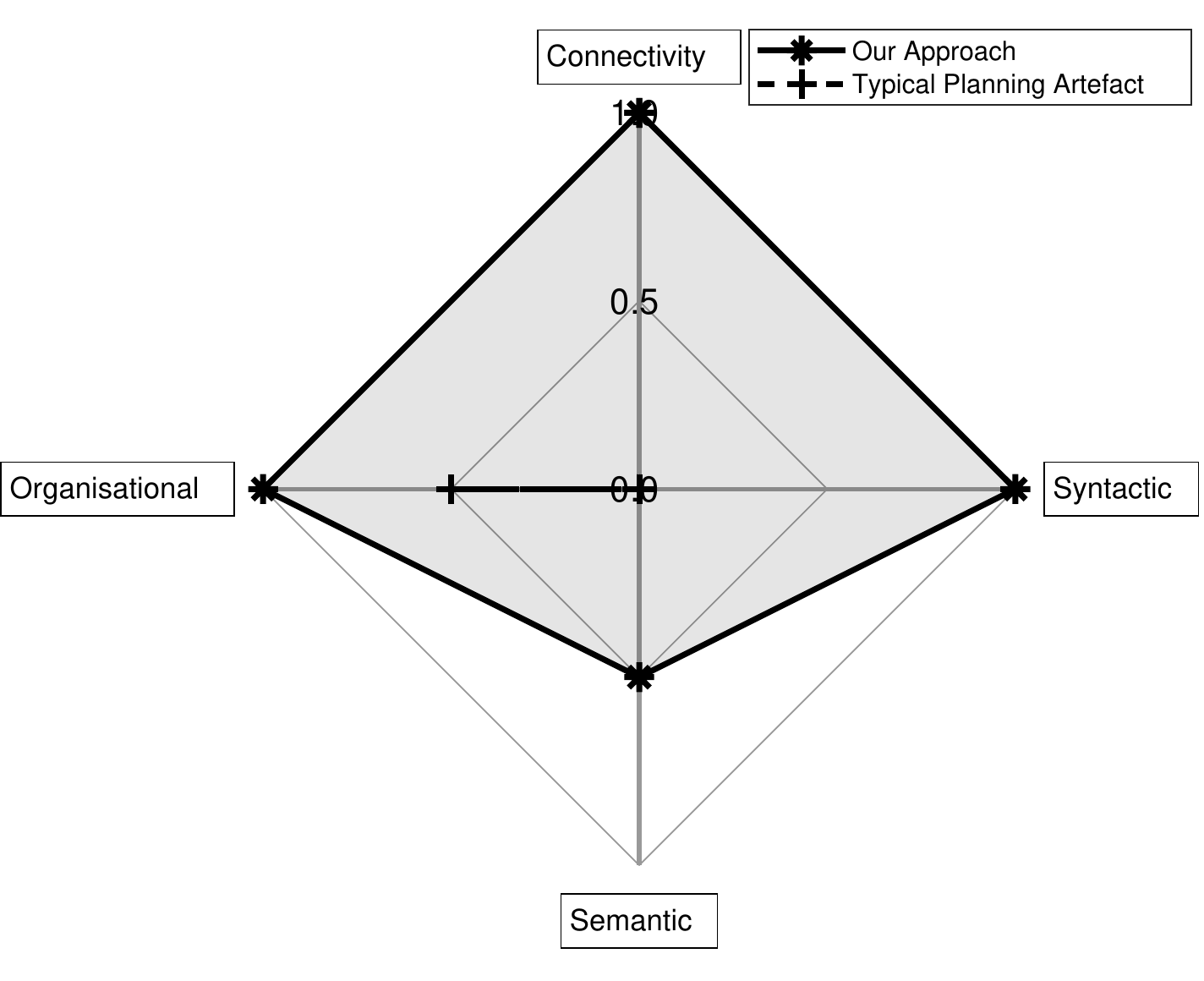}
  \caption{Radar Chart of the Interoperability Indicators}
  \label{fig:evaluation:interoperability:plot}
\end{figure}

We assess interoperability using its four levels as indicators: organisational, semantic, syntactic and connectivity level. \cref{fig:evaluation:interoperability:plot} shows the scores of the interoperability indicators. We can observe that our approach has better scores than a typical planning artefact. We discuss each indicator next.

\paragraph{Organisational}
Our approach achieves organisational interoperability guaranteed by the architectural design and the clarity of purpose of each capability. On the other hand, planning artefacts typically are not associated with architecture designs and the purpose of a certain artefact and its components should be deduced either from a relevant paper or code if available. As a consequence, our approach scores \enquote{$+$}, and a typical planning artefact scores \enquote{$O$}.

\paragraph{Semantic}
Due to the distribution of capabilities in our service-oriented architecture, API documentation is necessary, which we accordingly provide. Our approach gives processing guarantees by specifying the expected return type, but no mandatory validation is performed. Thus, it is theoretically possible to misinterpret objects if they are misused. This can be further reduced if service interfaces are sufficiently documented. Therefore, we assign the \enquote{$O$} score to our approach. In a typical planning artefact, semantics are only provided in a corresponding scientific paper, while software documentations are rarely given if at all. As a consequence, the score for a typical planning artefact is \enquote{$-$}. 

\paragraph{Syntactic}
Since syntactic interoperability requires an SOA, this indicator is not applicable to a typical planning artefact. So, the score is \enquote{$-$}. On the other hand, our approach is based on SOA and uses a uniform communication format (i.e., JSON). Thus, the score is \enquote{$+$}.

\paragraph{Connectivity}
Connectivity requires decentralisation, a property uncharacteristic for typical planning artefacts. We therefore assign the \enquote{$-$} score. On the other hand, our approach is decentralised and supported by commonly used standards and tools. For the front-end, RESTful is used, which is based on the HTTP protocol. The choice of MOM is RabbitMQ, which is based on the TCP protocol. Both protocols are supported in most programming languages. Finally, our approach does not restrict the choice of MOM. Thus, we award the \enquote{$+$} score.

\subsection{Reusability}
\begin{figure}
  \centering
  \includegraphics[width=\linewidth]{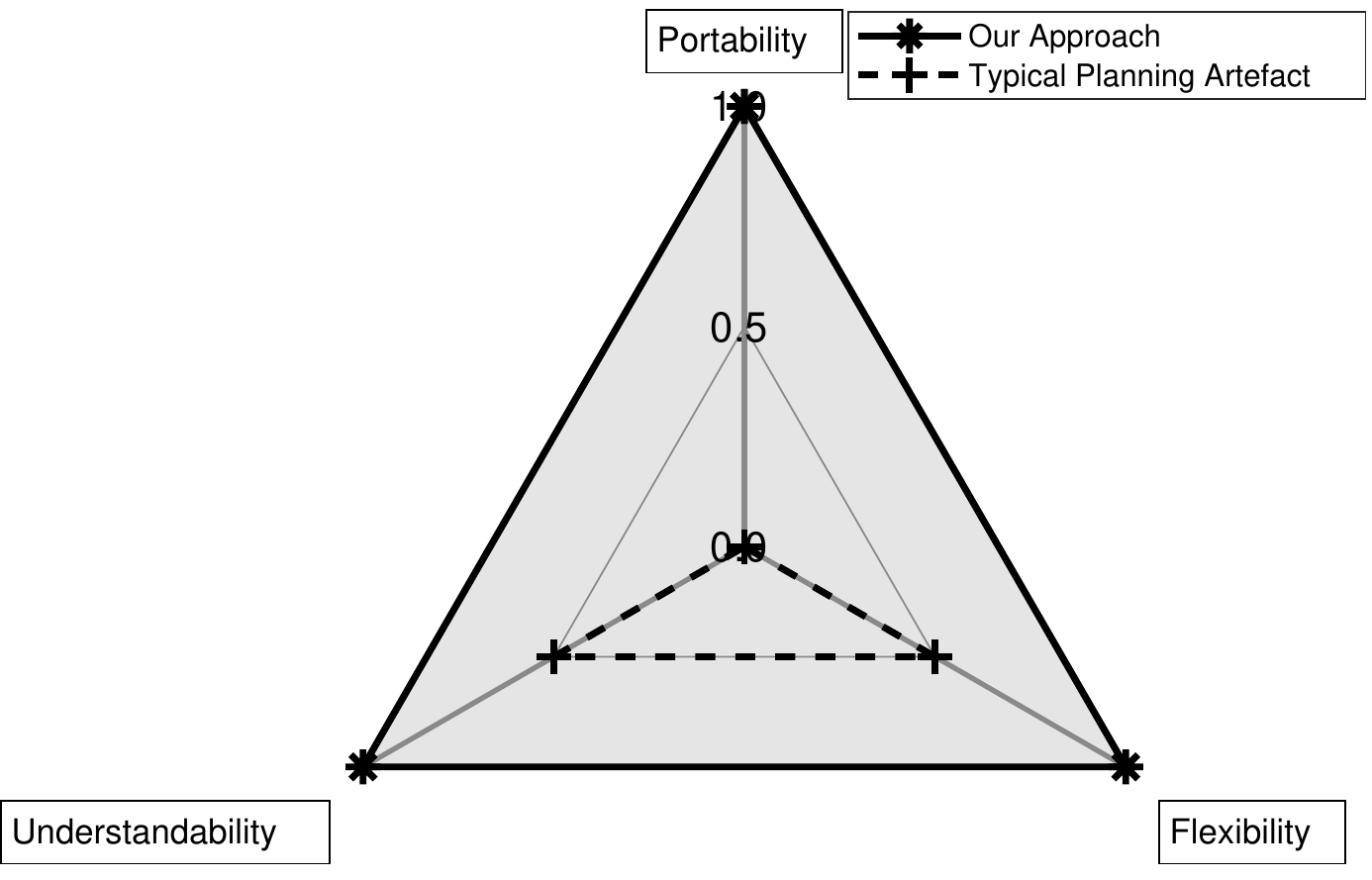}
  \caption{Radar Chart of Reusability Indicators}
  \label{fig:evaluation:reusability:plot}
\end{figure}

We assess reusability as a composition of the following indicators: portability, flexibility, and understandability~\cite{Cardino1997}. \cref{fig:evaluation:reusability:plot} shows the scores of the reusability indicators. It can be observed that our approach scores better in all indicators than a typical planning artefact. As before, we discuss each indicator next.

\paragraph{Portability}
Portability can be determined by having as few dependencies as possible. Our planning architecture uses services that encapsulate their dependencies. Although the services may have many dependencies to external libraries, the dependencies are handled during compile-time. It is also possible to use dependency handlers (e.g., Gradle, Maven) to minimise repository dependency. During runtime, the planning services are entirely decoupled from the communication interface. Therefore, we assign the \enquote{$+$} score to the proposed approach.

Most of existing planning artefacts (e.g., PandaPIParser~\cite{holler2021:panda}) depend on operating system libraries, which may be due to the programming language used (e.g., C++) and the design of the artefact's architecture. Rare are the planning artefacts that offer encapsulated solutions (e.g., virtual images). Even runnable artefacts are often challenging to find, so one has to handle all build dependencies. As a consequence, we assign \enquote{$-$} for the portability of a typical planning artefact.

\paragraph{Flexibility}
Flexibility consists of two components, namely generality and modularity. Our approach offers a high degree of generality and modularity. This is mainly due to the encapsulation of planning capabilities, which can be also interpreted as individual modules. Generality is also achieved by using standard communication. Finally, planning capabilites are general enough to compose them in various ways. Therefore, we assign the \enquote{$+$} score.

When examining existing planning artefacts, one can notice that most of them do not use any other modular structures apart from the division into packages. Also, the generality of most of the planning artefacts as software systems is low. However, a set of utility functions is often provided, which can be used independently from the application context. This allows a minimal degree of flexibility, thus, we assign the \enquote{$O$} score.

\paragraph{Understandability}
Understandability also consists of two components. The first one is documentation. Our approach may have detailed code documentation, however, there is no requirement that enforces such a description. Complexity is the second component. Our system's total complexity is distributed over several services, which decreases the functional complexity of the individual services. This should be taken with a grain of salt due to the communication effort that may incur in a distributed system. Since the functional complexity outweighs the technical complexity in our context, we assign the \enquote{$+$} score. 

For existing planning artefacts, the documentation is extremely rare. The behavior of the artefacts is typically presented in a scientific paper with no guarantee for consistency. Since typical planning artefacts bundle all their functionalities, their behavior is relatively complicated. Therefore, the comprehensibility of their individual components is impeded and worse than in a system with separate concerns. Therefore, we assign the \enquote{$O$} score.

\section{Discussion}\label{sec:discussion}
The collection of planning capabilities is a result of a qualitative analysis of a set of planning artefacts. This is performed in a rigorous and systematic way in which all steps and information are documented, guaranteeing its validity. While the set of analysed planning artefacts is not exhaustive, it represents a relevant sample of software tools, architectures and frameworks in which AI planning is the intervention. This ensures that the collection of planning capabilities can be generalisable to a wider population of planning artefacts. The validity and generalisability allow us to qualify our analysis and its results as trustworthy. 

Instead of a decentralised approach for the design our planning architecture, we can go for the option of a centralised approach that is realised through a central capability service for pure process control. Since the central capability service would need to be always adapted when changing a planning capability or the process itself, a centralised architecture would lack usability. In theory, such a service could be set up generically to be configured via its environment, but this would not be easy to maintain during production. Besides, scalability would be endangered.

Our planning architecture is designed around the concepts of loose coupling and messaging to allow for usability, interoperability and reusability of planning capabilities. This, however, does not entail specifications of standardised interfaces for the capabilities. The reason for this lies in the current state of AI planning and interest of the AI planning community. In particular, from a standardisation point of view, the community has so far focused only on developing standard modelling languages (PDDL for classical planning and HPDL/HDDL for hierarchical planning). As a consequence, no standards are available for planning capabilities, making it extremely difficult to design standard interfaces for planning capabilities such that will support the available planning tools. These findings are reflected in the planning architecture by allowing multiple instances per planning capability. This circumvents the need to operate all planning tools for a single planning capability via one interface. In addition, it is a reasonable approach to create JSON-based interfaces for the modeling languages. This opportunity is only possible for the interfaces of the Parsing capability.

Our planning architecture and prototype are based on messaging, which involves forwarding of objects and requires their classes to be serialisable. This means that the capabilities of existing planning artefacts need to be decomposable into separate serialisable classes. Existing planning capabilities are tightly coupled software systems, making it extremely difficult, almost impossible, to extract their capabilities as separate classes. This is mainly due to fact that the AI planning community has neglected software-engineering principles and focused exclusively on the behaviour of the planning tools and algorithms. So, the lack of serialisable classes significantly complicates the integration task. It is therefore necessary to integrate every existing planning implementation into a service using a wrapper with a serialisable data structure. We tried to do this for PDDL4J, JSHOP2~\cite{Ilghami2006DocumentationFJ}, KPlanning~\cite{krzisch} and PANDA, especially to extract the data structure for the Parsing capability. This proved also problematic, Many planners use grounding\footnote{Grounding consists of listing and instantiating all possible actions from the planning problem specification~\cite{Ramoul2017}.} directly linked to the parsing process. In some cases, we even got in touch with the system developers to clarify the problem of data interpretation during parsing. In the end, we successfully integrated only the functionality of the PDDL4J library. 

The Modelling service is implemented as a single front-end service. This makes the service complex and tightly coupled. A better approach would be to separate the front-end functionality into more discrete capabilities and decouple it from the functionality used to provide results, login and administration features. 

The high scores of usability, interoperability and reusability can be justified by the fact that the planning architecture is the first planning artefact that incorporates software-engineering principles, design patterns and practices. There are also scores lower than the maximum possible. For usability, SOA may cause some disadvantages, especially in planning requests. At the same time, SOA offers many possibilities to improve usability by allowing implementation of multiple capabilities. For interoperability, our approach does not enforce equal semantic treatment of interacting capabilities. The high interoperability increases the need to have stable or standard interfaces. A correct connection to error management is necessary to be able to create meaningful error messages. For reusability, the currently used JavaDocs might not be sufficient as documentation for such a platform planning architecture. Detailed documentation of each planning capability and corresponding interfaces might be necessary.. 

In addition to usability, interoperability and reusability, our approach has the advantage of flexibility. The planning architecture can be easily extended with other planning capabilities and also it can support any data model provided its serialisability. So, the proposed approach can be seen as an ecosystem of existing planning capabilities and incubator of new planning services, having the potential to avoid many development and integration problems and to save time and effort. These advantages would become even more apparent when the system gets connected to a real-world application.

\section{Conclusions}\label{sec:conclusion}
We put a software architecture at the core of the ability to build, sustain and foster the use of AI planning systems. We collected a set of planning capabilities that we classified from an operational and technical perspective. These two classifications provide initial support for designers and engineers of AI planning systems by allowing them to quickly understand the objectives, features and properties of the numerous planning capabilities. Upon these capabilities, we designed a service-oriented planning architecture that meets the requirements of usability, interoperability and reusability. The architecture design incorporates appropriate software-engineering principles and design patterns. 

The developed system prototype is used to demonstrate the potential for rapid prototyping by leveraging the flexibility of the planning architecture, but also the possibility to integrate existing planning tools when their complex internals allow for it. We showed that our proposed approach has qualitative attributes that go beyond those of typical AI planning artefacts. While our planning architecture represents an initial effort and the prototype offers only limited capabilities, we believe they make a significant step towards bringing closer software architecture and AI planning.

\bibliography{main.bib}

\end{document}